\documentclass[letterpaper]{article}

\usepackage{natbib,alifeconf}  
\usepackage{booktabs}
\usepackage{amsmath}
\usepackage{graphicx}
\usepackage{url,hyperref,cleveref}
\title{Mass Conservation as an Inductive Bias for\\Self-Organized Criticality in NCA Reservoirs}

\author{
    Tong Zhang$^{1}$,
    Etienne Guichard$^{2}$,
    Sidney Pontes-Filho$^{3}$, \and
    Stefano Nichele$^{2}$ \\
    $^1$Independent Researcher, Beijing, China \\
    $^2$\O stfold University College, Halden, Norway \\
    $^3$Simula Research Laboratory, Oslo, Norway \\
    stefano.nichele@hiof.no
}

\begin{document}

\maketitle

\begin{abstract}
Self-organized criticality (SOC), a dynamical regime associated with maximal information processing, offers a promising foundation for reservoir computing.
Recent work has shown that neural cellular automata (NCA) can be evolved toward critical avalanche dynamics and employed as effective reservoirs for memory and classification tasks.
Here, we investigate whether mass conservation---a local redistribution rule that preserves total lattice mass---serves as an inductive bias toward SOC in evolved NCA reservoirs.
We compare mass-conserving and standard NCA across multiple independent runs and evaluate both on three downstream benchmarks: 5-bit sequential memory, MNIST digit classification, and CartPole-v1 temporal control.
Mass-conserving NCA consistently exhibit stronger criticality, with more runs achieving perfect power-law fits across avalanche distributions, while also being 1.27$\times$ faster during evolution.
Importantly, conservation does not impair downstream utility: both variants achieve comparable performance across all three tasks.
Furthermore, the reservoir with perfect criticality achieves the highest temporal control score, suggesting a positive link between SOC quality and sequential computation.
Our results demonstrate that mass conservation is a simple, effective mechanism for promoting robust criticality in evolved NCA reservoirs without sacrificing downstream performance.
\end{abstract}

Data/Code available at: \url{https://github.com/modal905/mace-nca-reservoir}

\section{Introduction}

Self-organized criticality (SOC)~\citep{bak1987self} describes dynamical systems that spontaneously evolve toward a critical state characterized by scale-free avalanches dynamics and power-law statistics.
The edge-of-chaos hypothesis~\citep{langton1990computation} suggests that systems operating near the boundary between order and chaos can exhibit favorable computational properties, including enhanced information storage and transfer.
These ideas are particularly relevant to reservoir computing (RC)~\citep{jaeger2001echo,maass2002real}, which leverages the intrinsic dynamics of a fixed, high-dimensional system to map inputs into a feature space from which a simple linear readout can solve downstream tasks.

Cellular automata (CA) have long been studied as reservoir substrates~\citep{yilmaz2014reservoir,nichele2017neat}.
More recently, \citet{pontes2025reservoir} showed that neural cellular automata (NCA) evolved with CMA-ES~\citep{hansen2003reducing} toward critical avalanche statistics can function as effective reservoirs for memory and classification.
In parallel, \citet{papadopoulos2025mace} introduced MaCE, a mass-conserving update rule for cellular automata that redistributes activation locally while preserving total mass. This conservation constraint has been argued to regularize the dynamics by preventing runaway growth or collapse and by making complex behaviors more prevalent in the parameter space.

In this work, we ask whether mass conservation acts as an inductive bias toward SOC in evolved NCA reservoirs.
We incorporate a lightweight, MaCE-inspired conservation mechanism~\citep{papadopoulos2025mace} into the NCA architecture of \citet{pontes2025reservoir} and compare conserving and non-conserving variants across multiple independent evolutionary runs.
We evaluate both variants on three downstream tasks---5-bit sequential memory, MNIST classification, and CartPole temporal control---and analyze both their criticality and task performance.

Our contributions are:
\begin{enumerate}
    \item We show that mass conservation reliably produces better SOC: 2 out of 3 conserving seeds achieve perfect criticality (all six avalanche distributions pass the goodness-of-fit (GOF) test), while none out of 3 baseline seeds do.
    \item Conservation yields 7.3\% higher fitness and 1.27$\times$ faster training.
    \item Conservation does not sacrifice downstream performance: both variants achieve parity across all three tasks, and the conserving variant leads on temporal control after best-GOF checkpoint selection.
    \item The seed with the best criticality also achieves the best temporal task performance, suggesting a positive link between SOC quality and sequential computation.
\end{enumerate}

\section{Related Work}

\paragraph{Reservoir computing with cellular automata.}
Reservoir computing~\citep{jaeger2001echo, maass2002real} maps inputs through a high-dimensional dynamical system and trains only a linear readout, avoiding the cost of backpropagation through the reservoir. Cellular automata have been investigated as reservoirs since \citet{yilmaz2014reservoir} showed that certain elementary CA rules can perform computational tasks. \citet{nichele2017neat} extended this to deeper CA-based architectures, and \citet{glover2023investigating} provided a systematic analysis of CA reservoir rules and the 5-bit memory benchmark. \citet{glover2024when} further investigated when CA-based reservoirs outperform classical approaches in terms of downstream task performance.

\paragraph{Neural cellular automata.}
NCA extend classical CA by replacing hand-designed rules with learned neural network updates. \citet{mordvintsev2020growing} demonstrated that NCA can learn to grow and regenerate complex morphologies from a single cell, highlighting their capacity for robust, distributed computation. Subsequent work has applied NCA to self-organized control~\citep{variengien2021self} and critical pre-training for robotic locomotion~\citep{guichard2024critically}, establishing them as flexible substrates for a range of open-ended tasks beyond classification.

\paragraph{Evolving dynamical systems toward criticality.}
Several works have explored computational approaches to achieving criticality in distributed dynamical systems. \citet{pontesfilho2020evodynamic} introduced EvoDynamic, a general matrix-based framework for evolving CA and related systems toward critical behavior, enabling efficient optimization through deep learning libraries.  \citet{pontes2020neuro} proposed a neuro-inspired framework for evolving stochastic CA, random Boolean networks, and echo state networks toward power-law avalanche distributions. \citet{pontes2022assessing} subsequently assessed the robustness of critical behavior in stochastic CA, showing that evolved critical dynamics can withstand perturbations to initial conditions---a property closely related to self-organized criticality.

\paragraph{Critical NCA reservoirs.}
\citet{pontesfilho2023critical} first proposed evolving a deterministic, binary NCA toward critical avalanche statistics and outlined its potential as a reservoir computing substrate. This was fully realized by \citet{pontes2025reservoir}, who demonstrated that NCA evolved via CMA-ES~\citep{hansen2003reducing} can serve as effective reservoirs for memory and classification tasks, achieving MNIST accuracy on par with the best elementary CA rules. The edge-of-chaos hypothesis~\citep{langton1990computation, bertschinger2004real} provides the theoretical motivation: dynamical systems near a critical phase transition are predicted to exhibit maximal computational capacity.

\paragraph{Flow Lenia.}
\citet{plantec2023flowlenia} introduced mass-conserving dynamics into Lenia through a flow-based formulation, where mass is transported between cells according to locally computed velocity fields. This conservation constraint enables the emergence of self-organizing particle-like structures with lifelike behaviors, demonstrating that preserving total mass can enrich the behavioral repertoire of continuous cellular automata.

\paragraph{Mass-conserving cellular automata.}
The MaCE framework~\citep{papadopoulos2025mace} introduced a general mass-conserving evolution rule for CA based on local softmax-weighted redistribution. Conservation ensures that total lattice mass is preserved exactly at each timestep, preventing mass explosion or extinction. MaCE was shown to increase the density of interesting behaviors in parameter space and was demonstrated on Lenia, NCA, and discrete CA. Of particular relevance to our work, the MaCE-NCA variant applies conservation only to visible (RGB) channels while leaving hidden channels unconstrained---the same design choice we adopt. Together with Flow Lenia, MaCE demonstrates that conservation constraints can enrich the behavioral repertoire of CA; our work extends this principle to discrete, evolved NCA reservoirs.

\section{Methods}

\subsection{NCA Architecture}
Our NCA follows the architecture of \citet{pontes2025reservoir}: a one-dimensional, binary, deterministic cellular automaton with state updated by a small convolutional neural network. The NCA operates on a grid of width $W$ with $C = 5$ channels (1 visible + 4 hidden) using periodic boundary conditions. At each timestep, the full state tensor $\mathbf{s} \in \{0,1\}^{W \times C}$ is updated synchronously by a 3-layer Convolutional Neural Network (CNN):

\begin{table}[h]
\centering
\caption{NCA update network architecture (1,565 parameters).}
\label{tab:architecture}
\begin{tabular}{lcccc}
\hline
\textbf{Layer} & \textbf{Filters} & \textbf{Kernel} & \textbf{Activation} & \textbf{Params} \\
\hline
Conv1D    & 30 & 3 & ReLU      & 480 \\
Conv1D    & 30 & 1 & ReLU      & 930 \\
Conv1D    &  5 & 1 & threshold & 155 \\
\hline
\multicolumn{4}{l}{\textbf{Total}} & \textbf{1,565} \\
\hline
\end{tabular}
\end{table}

The first layer (kernel size 3) implements the local neighborhood rule; subsequent layers ($1 \times 1$ convolutions) act as pointwise nonlinearities. The final layer applies a Heaviside (step-like) threshold (positive numbers $\to$ $1$, nonpositive numbers $\to$ $0$) to produce the binary next state. All cells are updated simultaneously. This architecture is lightweight by design---small enough to be evolved with a derivative-free optimizer, yet sufficient to produce complex spatiotemporal dynamics.

\subsection{Mass-Conservation Mechanism}
Inspired by the MaCE framework of \citet{papadopoulos2025mace}, we add a conservation constraint to the NCA. Following the MaCE-NCA design, conservation is applied only to the visible channel while the four hidden channels remain unconstrained, allowing the neural network to develop arbitrary internal representations.

After each CNN update, the visible channel undergoes a local two-step \emph{donate-then-collect} redistribution that preserves total lattice mass exactly. Let $\hat{s}_i^{(0)}$ denote the raw (pre-conservation) visible-channel value at cell~$i$, with circular neighborhood $\mathcal{N}(i) = \{i{-}1,\, i,\, i{+}1\}$ (wrapped at boundaries). Each cell~$j$ first \textbf{donates} its mass proportionally to its neighbors via softmax-weighted affinity:
\begin{equation}
  d_{j \to i} \;=\;
  \frac{\exp(\hat{s}_i^{(0)})}
       {\displaystyle\sum_{k \in \mathcal{N}(j)} \exp(\hat{s}_k^{(0)})}
  \;\cdot\; \hat{s}_j^{(0)},
  \qquad i \in \mathcal{N}(j)
  \label{eq:donate}
\end{equation}
Then each cell~$i$ \textbf{collects} donations from all neighbors whose neighborhood includes~$i$:
\begin{equation}
  s_i^{(0)} \;=\; \sum_{j \in \mathcal{N}(i)} d_{j \to i}
  \label{eq:conservation}
\end{equation}
followed by the standard binarization threshold. In other words, conservation is applied to the continuous-valued visible-channel outputs (pre-binarization); the Heaviside threshold is then applied to obtain the binary visible state. Because every unit of mass donated by~$j$ is collected exactly once, the total visible activation $\sum_i s_i^{(0)}$ is preserved at each timestep. The 4 hidden channels remain unconstrained. The mechanism is differentiable in principle and adds zero trainable parameters.

\subsection{Evolutionary Training}
We evolve the 1,565 NCA weights using CMA-ES (Covariance Matrix Adaptation Evolution Strategy; \citealp{hansen2003reducing}), following the protocol of \citet{pontes2025reservoir}. CMA-ES maintains a population of 96 individuals per generation. Two variants are trained independently: a \emph{baseline} NCA (no conservation) and a \emph{conserving} NCA (Equation~\ref{eq:conservation} applied after each update). Training runs for 500 generations on a grid of width $W = 1000$ for $T = 1000$ timesteps, extending the original protocol of 100--200 generations at the same grid scale~\citep{pontes2025reservoir}.

The fitness function is a composite score rewarding power-law behavior in the visible channel's avalanche statistics, adapted from the framework of \citet{pontes2020neuro} and \citet{pontes2022assessing}. It combines the coefficient of determination ($R^2$) of least-squares regression in log-log space, the Kolmogorov--Smirnov (KS) distance between empirical and fitted power-law distributions, the percentage of non-zero bins in each distribution, the percentage of unique states visited during simulation, and log-likelihood ratios comparing power-law versus exponential models~\citep{clauset2009power}. These six component scores---one per avalanche distribution---are aggregated via sigmoid, exponential, and hyperbolic-tangent mappings into a single scalar fitness $S$ (see \citealp{pontes2025reservoir} for the full formulation). To ensure reproducibility, we train each variant independently using three distinct random initialization seeds (42, 43, 44) on AMD EPYC 9654 processors.

To ensure full reproducibility, the CMA-ES seed is supplied via the \texttt{--seed} command-line option (forwarded to \texttt{python-cma}) and logged to the run’s \texttt{args.json} file. The only other NumPy RNG reseed in the main training/evaluation path is the fixed initialization of the CA input via \texttt{np.random.seed(1)} inside \texttt{evaluate\_nca}. Thus exact replication requires both the same \texttt{--seed} and the same codebase/implementation.

\subsection{Criticality Evaluation}
Self-organized criticality is assessed via avalanche analysis on the visible channel. An avalanche is defined as a contiguous run of the same binary visible state across space and time: a 0-state avalanche is an inactive region bounded by active cells or boundaries, while a 1-state avalanche is an active region bounded by inactivity. For each state, we extract three distributions from the visible-channel timecourse: avalanche size (number of cells involved), duration (number of consecutive timesteps), and inter-arrival time (time between successive avalanche onsets); this yields six distributions in total.

For each distribution, we fit a power-law model using the \texttt{powerlaw} Python library \citep{clauset2009power} and assess goodness-of-fit by comparing the empirical KS distance to the distribution of KS distances obtained from 1,000 synthetic datasets sampled from the fitted model. The GOF test p-value is the fraction of synthetic distances that are larger than the empirical distance. A distribution passes if $p \geq 0.1$, meaning the empirical data are not statistically inconsistent with the fitted power law. This does not prove that the data uniquely follow a power law; it only indicates compatibility with the fitted model. The GOF test can also be insensitive for small or highly concentrated samples, in which case a steep distribution may pass "trivially." An NCA checkpoint achieves \emph{perfect criticality} if all six distributions pass.

\subsection{Downstream Tasks}
We evaluate the computational capability of evolved NCA reservoirs on three downstream tasks, following the reservoir computing paradigm \citep{jaeger2001echo, maass2002real}: the NCA dynamics serve as a fixed nonlinear expansion, and only a linear readout is trained.

\paragraph{5-bit sequential memory.}
Following \citet{pontes2025reservoir}, we test whether the reservoir retains a 5-bit input pattern after a 200-step distractor period. We use a smaller grid ($W = 80$, 4 input channels) and inject the 5-bit pattern at random positions. After 200 distractor steps, a linear SVM \citep{cortes1995support, pedregosa2011scikit} is trained to decode the original input from the flattened reservoir state. We run 100 independent experiments per seed; the score is the fraction of correctly classified trials (1.0 = perfect).

\paragraph{MNIST classification.}
Each MNIST image \citep{lecun1998mnist} is reshaped into a vector of size 784 and fed as initialization to the visible channel of a $W = 784$ NCA. After $T = 4$ timesteps, the final state is flattened into a feature vector. A LinearSVC with default scikit-learn hyperparameters~\citep{pedregosa2011scikit} is trained on the 60,000 training features and evaluated on the 10,000 test images, repeated 10 times per seed to estimate variance. This readout configuration follows \citet{pontes2025reservoir}, who used the same linear SVM approach.

\paragraph{CartPole-v1 temporal control.}
To test sequential decision-making, we use the CartPole-v1 environment \citep{brockman2016openai} with a reservoir of width $W = 100$ and $T = 100$ timesteps per action. Observations are one-hot encoded (20 bins per dimension, 4 dimensions) and injected at fixed positions. A linear Q-network ($\mathbf{W}_{\mathrm{out}} \cdot \mathbf{x} \to 2$ Q-values) is trained with Q-learning for 1,000 episodes and evaluated over 100 episodes with greedy action selection ($\varepsilon = 0$).

\section{Results}

\subsection{Evolutionary Training}

\begin{figure}[t]
\centering
\includegraphics[width=\columnwidth]{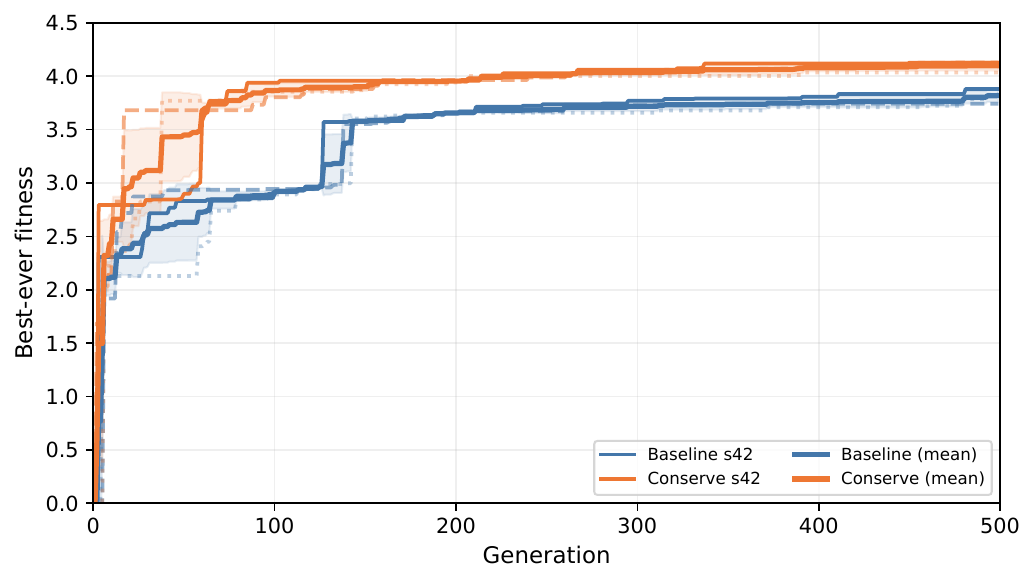}
\caption{Best-ever fitness trajectories over 500 generations for baseline (blue thick line) and conserving (orange thick line) variants across three seeds. Each thick line represents the mean fitness across three seeds, and the shaded region around it indicates $\pm$1 standard deviation. The thin lines within each band show individual seed trajectories.}
\label{fig:fitness}
\end{figure}

Table~\ref{tab:training} summarizes training outcomes across three independent seeds. The conserving variant achieves 7.3\% higher mean fitness (4.10 vs.\ 3.82) and converges 1.27$\times$ faster in wall-clock time (37.1\,h vs.\ 47.2\,h on average). Figure~\ref{fig:fitness} shows the best-ever fitness trajectories across all six training runs.

\begin{table}[h]
\centering
\caption{Training results across 3 seeds ($W\!=\!1000$, $T\!=\!1000$, 500 generations). GOF passes are from the best-fitness checkpoint during training.}
\label{tab:training}
\small
\begin{tabular*}{\columnwidth}{@{\extracolsep{\fill}}llcccc}
\hline
\textbf{Variant} & \textbf{Seed} & \textbf{Best Fit.} & \textbf{Best Gen.} & \textbf{Time (h)} & \textbf{GOF} \\
\hline
Baseline  & 42 & 3.882 & 481 & 46.6 & 3/6 \\
           & 43 & 3.743 & 349 & 42.5 & 2/6 \\
           & 44 & 3.832 & 499 & 52.5 & 2/6 \\
           & \textbf{Avg} & \textbf{3.82} & --- & \textbf{47.2} & \textbf{2.3/6} \\
\hline
Conserve  & 42 & 4.128 & 499 & 39.7 & 3/6 \\
           & 43 & 4.129 & 499 & 37.2 & \textbf{6/6} \\
           & 44 & 4.036 & 499 & 34.3 & \textbf{6/6} \\
           & \textbf{Avg} & \textbf{4.10} & --- & \textbf{37.1} & \textbf{5.0/6} \\
\hline
\end{tabular*}
\end{table}

\subsection{SOC Characterization}

Table~\ref{tab:soc} reports the post-training goodness-of-fit $p$-values for each of the six avalanche distributions (0-state and 1-state size, duration, and inter-arrival time). A distribution passes the GOF test if $p \geq 0.1$.

\begin{table}[h]
\centering
\caption{Post-training GOF $p$-values for 6 avalanche distributions. $p_1$--$p_3$: 0-state size, duration, and interarrival; $p_4$--$p_6$: 1-state size, duration, and interarrival. A distribution passes if $p \geq 0.1$. Bold indicates perfect criticality (6/6 passes).}
\label{tab:soc}
\small
\begin{tabular*}{\columnwidth}{@{\extracolsep{\fill}}llccccccc}
\hline
\textbf{Seed} & \textbf{Variant} & $p_1$ & $p_2$ & $p_3$ & $p_4$ & $p_5$ & $p_6$ & \textbf{Pass} \\
\hline
42 & Baseline & 1.0 & 1.0 & 0.36 & 0.0 & 0.0 & 0.0 & 3/6 \\
42 & Conserve & 0.0 & 0.0 & 0.0 & 1.0 & 1.0 & 1.0 & 3/6 \\
43 & Baseline & 0.0 & 0.0 & 0.0 & 1.0 & 1.0 & 1.0 & 3/6 \\
43 & Conserve & \textbf{1.0} & \textbf{1.0} & \textbf{1.0} & \textbf{1.0} & \textbf{1.0} & \textbf{1.0} & \textbf{6/6} \\
44 & Baseline & 1.0 & 0.0 & 1.0 & 1.0 & 1.0 & 1.0 & 5/6 \\
44 & Conserve & \textbf{1.0} & \textbf{1.0} & \textbf{1.0} & \textbf{1.0} & \textbf{1.0} & \textbf{1.0} & \textbf{6/6} \\
\hline
\end{tabular*}
\end{table}

\begin{figure}[t]
\centering
\includegraphics[width=0.85\columnwidth]{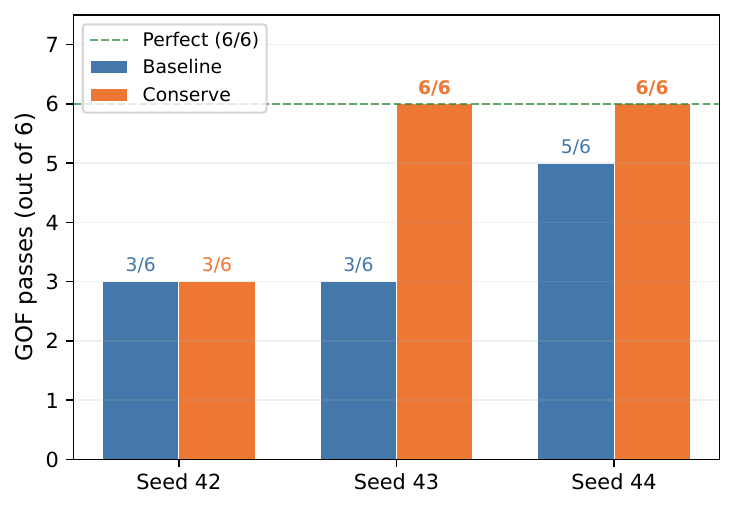}
\caption{Goodness-of-fit passes (out of 6 avalanche distributions) per seed at the gen-499 checkpoint. Conserving seeds 43 and 44 achieve perfect criticality (6/6); no baseline seed does.}
\label{fig:gof}
\end{figure}

\begin{figure*}[t]
\centering
\includegraphics[width=\textwidth]{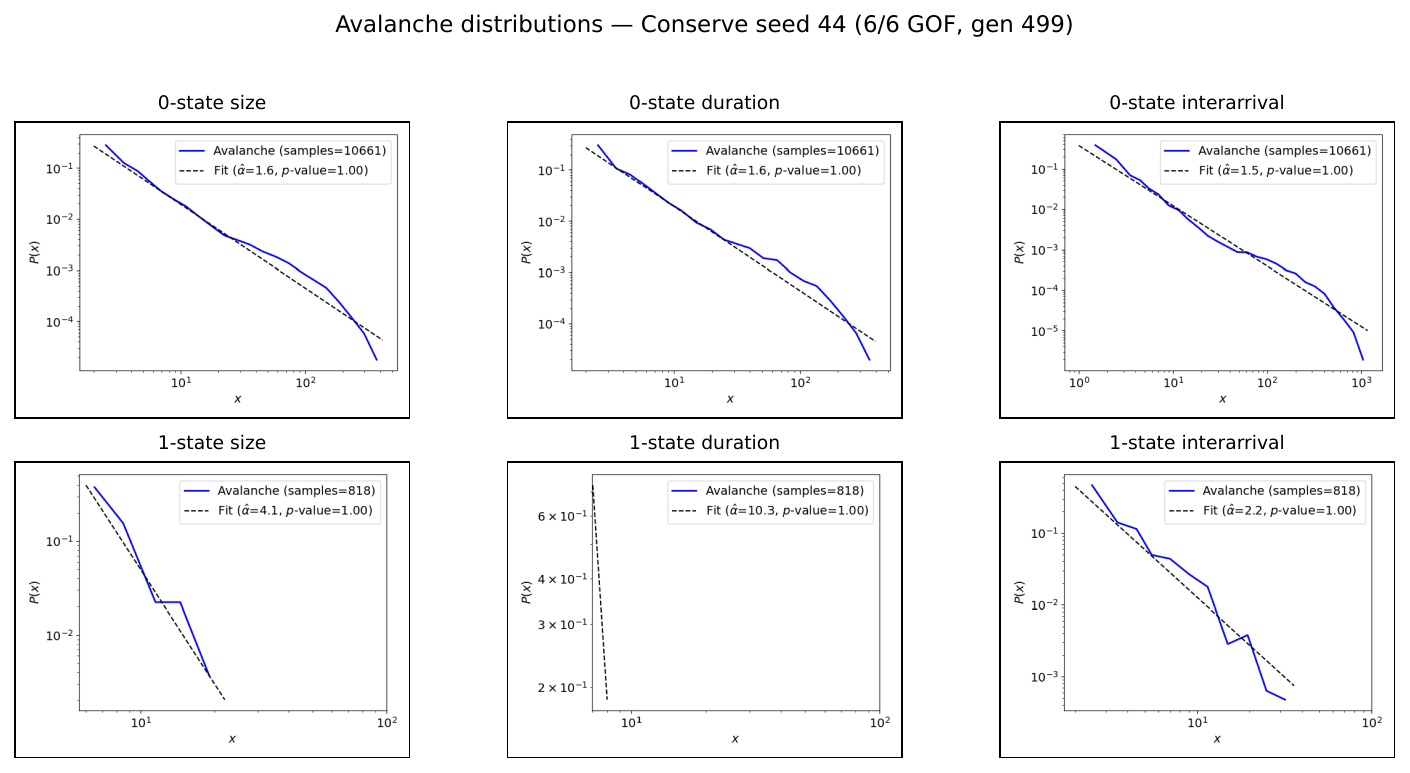}
\caption{Avalanche size, duration, and interarrival distributions for conserve seed~44 (6/6 GOF) at gen~499, arranged in a 2×3 grid: 0-state (top row: size, duration, interarrival) and 1-state (bottom row: size, duration, interarrival). All six distributions pass the GOF test ($p = 1.0$). The 1-state duration (bottom center) has a steep exponent ($\hat{\alpha} = 10.3$), indicating that most 1-state avalanches are very short-lived. This distribution passes the GOF test trivially because the concentrated distribution of durations has low variance, making the empirical KS distance small by chance; as discussed in the Criticality Evaluation section, this does not necessarily indicate robust power-law behavior.}
\label{fig:avalanche}
\end{figure*}

Only conserving seeds 43 and 44 achieve perfect criticality---all six avalanche distributions exhibit statistically significant power-law scaling ($p = 1.0$). No baseline seed achieves 6/6. Notably, both variants achieve at least 3/6 passes at every seed, indicating that CMA-ES evolution toward critical fitness is effective in both cases, but conservation provides a stronger inductive bias that pushes additional distributions into the critical regime. Figure~\ref{fig:gof} visualizes the GOF passes per seed. Figure~\ref{fig:avalanche} shows the six avalanche distributions for conserve seed~44: the 0-state distributions (size, duration, interarrival; top row) each span two orders of magnitude with exponents $\hat{\alpha} \approx 1.5$--$1.6$, while the 1-state size and interarrival distributions follow similar power laws ($\hat{\alpha} \approx 2.2$--$4.1$). The 1-state duration is an exception---nearly all 818 avalanches last only a few timesteps, yielding a steep exponent ($\hat{\alpha} = 10.3$) that passes the GOF test trivially.

\subsection{5-bit Sequential Memory}

Both variants achieve perfect memory scores across all seeds (Table~\ref{tab:5bit}). This confirms that the NCA reservoir, at width $W = 80$ with a 200-step distractor, retains sufficient sequential memory capacity regardless of whether conservation is applied.

\begin{table}[h]
\centering
\caption{5-bit memory scores (100 experiments per seed).}
\label{tab:5bit}
\begin{tabular}{lcc}
\hline
\textbf{Seed} & \textbf{Baseline} & \textbf{Conserve} \\
\hline
42 & 1.0 & 1.0 \\
43 & 1.0 & 1.0 \\
44 & 1.0 & 1.0 \\
\hline
\end{tabular}
\end{table}

\subsection{MNIST Classification}

Table~\ref{tab:mnist} reports MNIST test accuracy using LinearSVC readouts. Both NCA variants substantially outperform the raw-pixel baseline (91.12\%), confirming that the reservoir dynamics provide useful nonlinear features. The baseline NCA leads by a small margin (+0.27 percentage points on average), well within seed-to-seed variance, indicating task parity between variants.

\begin{table}[h]
\centering
\caption{MNIST test accuracy (\%) with LinearSVC readout (10 runs per seed). Raw-pixel baseline: 91.12\%. $\dagger$ = best-GOF checkpoint.}
\label{tab:mnist}
\begin{tabular}{lccc}
\hline
\textbf{Seed} & \textbf{Baseline} & \textbf{Conserve} & $\Delta$ \\
\hline
42 & 93.62 $\pm$ 0.13 & 93.18 $\pm$ 0.07 & +0.44 \\
43$^\dagger$ & 94.11 $\pm$ 0.18 & 93.75 $\pm$ 0.09 & +0.36 \\
44$^\dagger$ & 93.71 $\pm$ 0.16 & 93.69 $\pm$ 0.16 & +0.02 \\
\textbf{Avg} & \textbf{93.81 $\pm$ 0.16} & \textbf{93.54 $\pm$ 0.11} & \textbf{+0.27} \\
\hline
\end{tabular}
\end{table}

Conserving reservoir features are also 2--6$\times$ faster to classify (SVM fit time: 533--1880\,s vs.\ 3585--5630\,s), suggesting better linear separability of the conserving reservoir's feature representations.

\subsection{CartPole Temporal Control}

Table~\ref{tab:cartpole} reports mean reward over 100 evaluation episodes with greedy action selection. Neither variant reaches the CartPole-v1 ``solved'' threshold of 195, which is expected given the minimal architecture (1,565-parameter NCA with a linear Q-network). However, both variants achieve control well above random ($\approx$20), and individual episodes reach the 500-step cap.

\begin{table}[h]
\centering
\caption{CartPole-v1 mean reward over 100 evaluation episodes. $\dagger$ = best-GOF checkpoint.}
\label{tab:cartpole}
\begin{tabular}{lccc}
\hline
\textbf{Seed} & \textbf{Baseline} & \textbf{Conserve} & \textbf{Winner} \\
\hline
42 & 100.7 $\pm$ 51.2  & 57.7 $\pm$ 85.4  & Baseline \\
43$^\dagger$ & 87.8 $\pm$ 66.4  & 140.7 $\pm$ 89.6 & Conserve \\
44$^\dagger$ & 77.3 $\pm$ 53.0  & \textbf{163.4 $\pm$ 97.3} & Conserve \\
\textbf{Avg} & \textbf{88.6} & \textbf{120.6} & Conserve \\
\hline
\end{tabular}
\end{table}

The conserving variant leads on average (120.6 vs.\ 88.6), and conserve seed~44---which also has the best criticality (6/6 GOF)---achieves the highest individual score (163.4). Baseline seeds~43 and~44 use best-GOF checkpoints (gen~201 and gen~492, respectively); their gen-499 scores were higher (159.2 and 146.7), revealing that checkpoint selection for criticality reduces temporal task performance in the baseline but not in the conserving variant. The high variance across seeds warrants further investigation with more seeds.

\begin{figure}[t]
\centering
\includegraphics[width=\columnwidth]{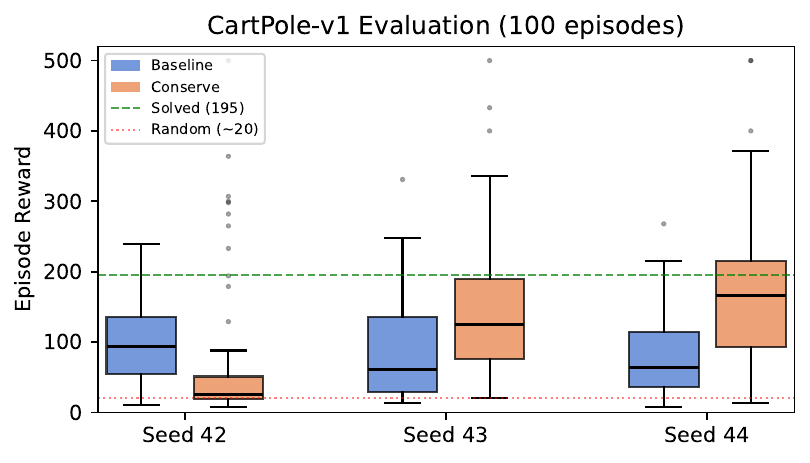}
\caption{CartPole-v1 evaluation reward distributions (100 episodes per seed). The green dashed line marks the solved threshold (195); the red dotted line marks random performance ($\approx$20). Conserve seed~44 (6/6 GOF) achieves the highest median reward.}
\label{fig:cartpole}
\end{figure}

\begin{figure*}[t]
\centering
\includegraphics[width=\textwidth]{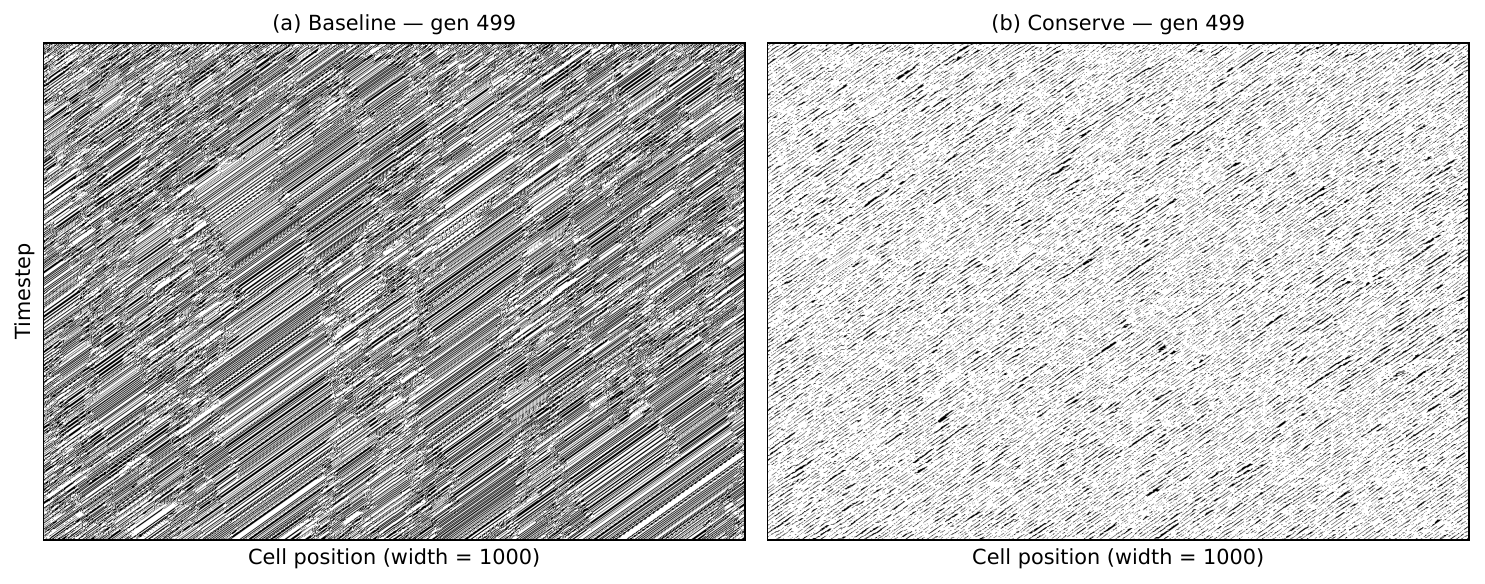}
\caption{Spacetime diagrams of the visible channel (1000 cells $\times$ 1000 timesteps) for (a) baseline and (b) conserving NCA at gen~499, seed~44. Black = active, white = inactive.}
\label{fig:spacetime}
\end{figure*}

Figure~\ref{fig:cartpole} shows the full distribution of episode rewards across seeds, and Figure~\ref{fig:spacetime} shows representative spacetime diagrams of the visible channel for baseline and conserving NCA at gen~499, illustrating the qualitative differences in spatiotemporal dynamics.

\section{Discussion}

\paragraph{Conservation as an inductive bias for SOC.}
Our central finding is that visible-channel mass conservation acts as a reliable inductive bias toward achieving visible-channel avalanche statistics that pass self-organized criticality tests. While CMA-ES evolves both variants toward critical fitness, conservation provides an additional structural constraint that channels the search toward configurations where \emph{all} avalanche distributions---not just a subset---exhibit power-law scaling. Two of three conserving seeds achieve perfect criticality (6/6 GOF), whereas no baseline seed does, despite comparable or even high partial scores (e.g., baseline seed~44 reaches 5/6). This suggests that conservation narrows the fitness landscape around fully critical attractors, making them easier for the optimizer to find.

\paragraph{Training efficiency.}
The 7.3\% fitness advantage and 1.27$\times$ wall-clock speedup of the conserving variant are consistent with the hypothesis that conservation regularizes the dynamics: by preventing mass explosion or extinction, the optimizer encounters fewer degenerate configurations and converges faster. This echoes findings in the MaCE framework~\citep{papadopoulos2025mace}, where conservation was observed to increase the density of interesting behaviors in parameter space.

\paragraph{Downstream task parity.}
A key concern when adding constraints to a reservoir is whether they reduce computational expressiveness. Our results indicate that conservation does not sacrifice downstream performance. Both variants achieve perfect 5-bit memory, near-identical MNIST accuracy ($<$0.3 percentage point difference), and CartPole control where the conserving variant leads on average (120.6 vs.\ 88.6) after best-GOF checkpoint selection. The faster SVM fit times for conserving features further suggest that conservation produces more linearly separable representations, which may be advantageous for tasks requiring simple readouts.

\paragraph{Criticality--computation link.}
The conserving variant leads on CartPole on average (120.6 vs.\ 88.6), and conserve seed~44---the seed with the best criticality (6/6 GOF)---achieves the highest individual score (163.4), suggesting a positive relationship between SOC quality and temporal computational capability. This aligns with theoretical arguments that critical systems maximize information transmission~\citep{langton1990computation, bertschinger2004real}, with the robustness analysis of critical CA dynamics by \citet{pontes2022assessing}, and with empirical observations in CA-based reservoirs~\citep{pontes2025reservoir}. However, with only three seeds per variant, this correlation remains preliminary and requires validation with larger seed populations.

\paragraph{Criticality--utility alignment.}
A notable asymmetry emerges when comparing checkpoint selection strategies. For the conserving variant, the final checkpoint (gen~499) simultaneously achieves the highest fitness and perfect criticality---no re-selection is needed. For the baseline, checkpoints with the best avalanche statistics do not coincide with those producing the best downstream performance. This suggests that conservation does not merely make criticality easier to achieve; it makes criticality \emph{compatible with functional performance}, eliminating the tension between SOC quality and task utility.

\paragraph{Limitations.}
This study has several limitations. First, the use of only three independent seeds per variant limits statistical power; future work should include 10 or more seeds for robust validation. Second, the CartPole task employs a deliberately minimal architecture (a linear Q-network), and its failure to reach the solved threshold (a reward of 195) stems from this design choice rather than any inherent limitation of the NCA reservoir. Third, conservation is applied only to the visible channel, and extending it to hidden channels may yield different dynamics. Finally, the current fitness function prioritizes criticality over direct optimization for downstream task performance—co-optimization could potentially produce reservoirs that are both critical and task-specialized.

\section{Conclusion}

We showed that mass conservation provides a reliable inductive bias toward self-organized criticality in evolved neural cellular automata reservoirs.
Across three independent seeds at scale ($W\!=\!1000$, $T\!=\!1000$, 500 generations), mass-conserving NCA exhibited substantially stronger power-law avalanche statistics than standard NCA: 2/3 conserving seeds passed all 6 goodness-of-fit tests, compared to 0/3 baseline seeds.
Conservation also improved optimization, yielding 7.3\% higher fitness and 1.27$\times$ faster convergence during evolution, while preserving downstream task performance on 5-bit sequential memory (perfect), MNIST classification ($\sim$93.5--93.8\%), and CartPole-v1 temporal control.

These results establish mass conservation as a lightweight, zero-parameter mechanism for promoting robust criticality in NCA reservoirs without sacrificing downstream performance. The seed exhibiting the strongest criticality also achieved the best temporal control performance, providing tentative support for the edge-of-chaos hypothesis in reservoir computing~\citep{langton1990computation, bertschinger2004real}. Future work should expand the seed population, explore conservation on hidden channels---as demonstrated for RGB channels in MaCE-NCA~\citep{papadopoulos2025mace}---and investigate the co-optimization of criticality and task fitness using frameworks such as EvoDynamic~\citep{pontesfilho2020evodynamic}.

\footnotesize
\bibliographystyle{apalike}
\bibliography{references}

@inproceedings{pontes2025reservoir,
  author    = {Pontes-Filho, Sidney and Nichele, Stefano and Lepperød, Mikkel},
  title     = {Reservoir Computing with Evolved Critical Neural Cellular Automata},
  booktitle = {Artificial Life Conference Proceedings},
  volume    = {37},
  pages     = {4},
  year      = {2025},
  publisher = {MIT Press}
}

@inproceedings{papadopoulos2025mace,
  author    = {Papadopoulos, Vassilis and Guichard, Etienne},
  title     = {{MaCE}: General Mass Conserving Dynamics for {CA}s},
  booktitle = {Artificial Life Conference Proceedings},
  volume    = {37},
  pages     = {18},
  year      = {2025},
  publisher = {MIT Press}
}

@article{bak1987self,
  author    = {Bak, Per and Tang, Chao and Wiesenfeld, Kurt},
  title     = {Self-organized criticality: An explanation of the 1/f noise},
  journal   = {Physical Review Letters},
  volume    = {59},
  number    = {4},
  pages     = {381--384},
  year      = {1987}
}

@article{langton1990computation,
  author    = {Langton, Christopher G.},
  title     = {Computation at the edge of chaos: Phase transitions and emergent computation},
  journal   = {Physica D: Nonlinear Phenomena},
  volume    = {42},
  number    = {1-3},
  pages     = {12--37},
  year      = {1990}
}

@article{jaeger2001echo,
  author    = {Jaeger, Herbert},
  title     = {The ``echo state'' approach to analysing and training recurrent neural networks},
  institution = {German National Research Center for Information Technology},
  journal   = {GMD Technical Report},
  volume    = {148},
  year      = {2001}
}

@article{maass2002real,
  author    = {Maass, Wolfgang and Natschl\"{a}ger, Thomas and Markram, Henry},
  title     = {Real-time computing without stable states: A new framework for neural computation based on perturbations},
  journal   = {Neural Computation},
  volume    = {14},
  number    = {11},
  pages     = {2531--2560},
  year      = {2002}
}

@article{yilmaz2014reservoir,
  author    = {Yilmaz, Ozg\"{u}r},
  title     = {Reservoir computing using cellular automata},
  journal   = {arXiv preprint arXiv:1410.0162},
  year      = {2014}
}

@article{hansen2003reducing,
  author    = {Hansen, Nikolaus and Ostermeier, Andreas},
  title     = {Completely derandomized self-adaptation in evolution strategies},
  journal   = {Evolutionary Computation},
  volume    = {9},
  number    = {2},
  pages     = {159--195},
  year      = {2001}
}

@article{clauset2009power,
  author    = {Clauset, Aaron and Shalizi, Cosma Rohilla and Newman, Mark E. J.},
  title     = {Power-law distributions in empirical data},
  journal   = {SIAM Review},
  volume    = {51},
  number    = {4},
  pages     = {661--703},
  year      = {2009}
}

@article{bertschinger2004real,
  author    = {Bertschinger, Nils and Natschl\"{a}ger, Thomas},
  title     = {Real-time computation at the edge of chaos in recurrent neural networks},
  journal   = {Neural Computation},
  volume    = {16},
  number    = {7},
  pages     = {1413--1436},
  year      = {2004}
}

@article{lecun1998mnist,
  author    = {LeCun, Yann and Bottou, L\'{e}on and Bengio, Yoshua and Haffner, Patrick},
  title     = {Gradient-based learning applied to document recognition},
  journal   = {Proceedings of the IEEE},
  volume    = {86},
  number    = {11},
  pages     = {2278--2324},
  year      = {1998}
}

@article{brockman2016openai,
  author    = {Brockman, Greg and Cheung, Vicki and Pettersson, Ludwig and Schneider, Jonas and Schulman, John and Tang, Jie and Zaremba, Wojciech},
  title     = {{OpenAI Gym}},
  journal   = {arXiv preprint arXiv:1606.01540},
  year      = {2016}
}

@article{nichele2017neat,
  author    = {Nichele, Stefano and Molund, Andreas},
  title     = {Deep learning with cellular automaton-based reservoir computing},
  journal   = {Complex Systems},
  volume    = {26},
  number    = {4},
  pages     = {319--340},
  year      = {2017}
}

@inproceedings{mordvintsev2020growing,
  author    = {Mordvintsev, Alexander and Randazzo, Ettore and Niklasson, Eyvind and Levin, Michael},
  title     = {Growing Neural Cellular Automata},
  booktitle = {Distill},
  year      = {2020},
  doi       = {10.23915/distill.00023}
}

@article{glover2023investigating,
  author    = {Glover, Thomas E. and Lind, Pedro and Yazidi, Anis and Osipov, Evgenii and Nichele, Stefano},
  title     = {Investigating rules and parameters of reservoir computing with elementary cellular automata, with a criticism of rule 90 and the five-bit memory benchmark},
  journal   = {Complex Systems},
  volume    = {32},
  number    = {3},
  pages     = {309--351},
  year      = {2023}
}

@article{glover2024when,
  author    = {Glover, Thomas and Osipov, Evgenii and Nichele, Stefano},
  title     = {On when is reservoir computing with cellular automata beneficial?},
  journal   = {arXiv preprint arXiv:2407.09501},
  year      = {2024}
}

@article{cortes1995support,
  author    = {Cortes, Corinna and Vapnik, Vladimir},
  title     = {Support-vector networks},
  journal   = {Machine Learning},
  volume    = {20},
  pages     = {273--297},
  year      = {1995}
}

@article{pontes2020neuro,
  author    = {Pontes-Filho, Sidney and Lind, Pedro and Yazidi, Anis and Zhang, Jianhua and Hammer, Hugo and Mello, Gustavo B. and Sandvig, Ioanna and Tufte, Gunnar and Nichele, Stefano},
  title     = {A neuro-inspired general framework for the evolution of stochastic dynamical systems: Cellular automata, random boolean networks and echo state networks towards criticality},
  journal   = {Cognitive Neurodynamics},
  volume    = {14},
  number    = {5},
  pages     = {657--674},
  year      = {2020}
}

@article{pedregosa2011scikit,
  author    = {Pedregosa, Fabian and Varoquaux, Ga\"{e}l and Gramfort, Alexandre and Michel, Vincent and Thirion, Bertrand and Grisel, Olivier and Blondel, Mathieu and Prettenhofer, Peter and Weiss, Ron and Dubourg, Vincent and others},
  title     = {Scikit-learn: Machine learning in {Python}},
  journal   = {Journal of Machine Learning Research},
  volume    = {12},
  pages     = {2825--2830},
  year      = {2011}
}

@inproceedings{variengien2021self,
  author    = {Variengien, Alexandre and Nichele, Stefano and Glover, Thomas and Pontes-Filho, Sidney},
  title     = {Towards self-organized control: Using neural cellular automata to robustly control a cart-pole agent},
  booktitle = {Innovations in Machine Intelligence},
  volume    = {1},
  pages     = {1--14},
  year      = {2021}
}

@mastersthesis{guichard2024critically,
  author    = {Guichard, Etienne},
  title     = {Critically pre-trained neural cellular automata as robot controllers},
  school    = {Delft University of Technology},
  year      = {2024}
}

@article{pontes2022assessing,
  author    = {Pontes-Filho, Sidney and Lind, Pedro G. and Nichele, Stefano},
  title     = {Assessing the robustness of critical behavior in stochastic cellular automata},
  journal   = {Physica D: Nonlinear Phenomena},
  volume    = {441},
  pages     = {133507},
  year      = {2022}
}

@inproceedings{pontesfilho2023critical,
  author    = {Pontes-Filho, Sidney and Nichele, Stefano and Lepperød, Mikkel},
  title     = {Critical neural cellular automata},
  booktitle = {ALIFE 2023: Ghost in the Machine Workshop},
  year      = {2023}
}

@inproceedings{pontesfilho2020evodynamic,
  author    = {Pontes-Filho, Sidney and Lind, Pedro and Yazidi, Anis and Zhang, Jianhua and Hammer, Hugo and Mello, Gustavo B. M. and Sandvig, Ioanna and Tufte, Gunnar and Nichele, Stefano},
  title     = {{EvoDynamic}: A framework for the evolution of generally represented dynamical systems and its application to criticality},
  booktitle = {International Conference on the Applications of Evolutionary Computation (EvoStar)},
  pages     = {133--148},
  publisher = {Springer},
  year      = {2020}
}

@inproceedings{plantec2023flowlenia,
  author    = {Plantec, Erwan and Hamon, Gautier and Etcheverry, Mayalen and Oudeyer, Pierre-Yves and Moulin-Frier, Cl\'{e}ment and Chan, Bert Wang-Chak},
  title     = {Flow {L}enia: Mass conservation for the study of virtual creatures in continuous cellular automata},
  booktitle = {Companion Proceedings of the Conference on Artificial Life (ALIFE)},
  year      = {2023}
}

\end{document}